\pdfoutput=1

\documentclass[11pt]{article}

\usepackage[preprint]{acl}

\usepackage{times}
\usepackage{latexsym}
\usepackage{booktabs}

\usepackage[T1]{fontenc}

\usepackage[utf8]{inputenc}

\usepackage{microtype}

\usepackage{inconsolata}

\usepackage{graphicx}
\usepackage{xspace}

\newcommand{\DisambigModel}{\texttt{LinkWikipedia}\xspace}
\newcommand{\DisambigNewsModel}{\texttt{LinkNewsWikipedia}\xspace}
\newcommand{\CorefModel}{\texttt{LinkMentions}\xspace}
\newcommand{\EoTU}{\texttt{Entities of the Union}\xspace}

\newcommand{\WikiDataset}{\texttt{WikiConfusables}\xspace}
\newcommand{\NewsHNDataset}{\texttt{NewsConfusables}\xspace}

\title{Contrastive Entity Coreference and Disambiguation for Historical Texts}

\author
{Abhishek Arora$^{1\dagger}$, Emily Silcock$^{1\dagger}$,  Leander Heldring$^{3}$, Melissa Dell$^{1, 2\ast}$ \\
\normalsize{$^{1}$Harvard University; Cambridge, MA, USA.}\\
\normalsize{$^{2}$National Bureau of Economic Research; Cambridge, MA, USA.}\\
\normalsize{$^{3}$Kellogg School of Management, Northwestern University, Evanston, IL, USA.}\\
\normalsize{$^{\dagger}$ These authors contributed equally.}
\normalsize{$^{\ast}$Corresponding author:  melissadell@fas.harvard.edu.}
}

\begin{document}
\maketitle
\begin{abstract}
Massive-scale historical document collections are crucial for social science research. Despite increasing digitization, these documents typically lack unique cross-document identifiers for individuals mentioned within the texts, as well as individual identifiers from external knowledgebases like Wikipedia/Wikidata. Existing entity disambiguation methods often fall short in accuracy for historical documents, which are replete with individuals not remembered in contemporary knowledgebases. This study makes three key contributions to improve cross-document coreference resolution and disambiguation in historical texts: a massive-scale training dataset replete with hard negatives - that sources over 190 million entity pairs from Wikipedia contexts and disambiguation pages - high-quality evaluation data from hand-labeled historical newswire articles, and trained models evaluated on this historical benchmark. We contrastively train bi-encoder models for coreferencing and disambiguating individuals in historical texts, achieving accurate, scalable performance that identifies out-of-knowledgebase individuals.  Our approach significantly surpasses other entity disambiguation models on our historical newswire benchmark. Our models also demonstrate competitive performance on modern entity disambiguation benchmarks, particularly certain news disambiguation datasets.

\end{abstract}

\section{Introduction}
\label{sec:intro}

Massive scale historical document collections - such as historical newspapers or the 14 billion documents in the U.S. National Archives - are central source materials for social science research. While historical documents are increasingly being digitized, they are not typically tagged with unique cross-document identifiers for individuals mentioned in the texts,  or with individual identifiers from an external knowledgebase such as English Wikipedia/Wikidata, which provides structured data for over a million individuals. 

While there is a large literature on entity disambiguation to a knowledgebase (but to a lesser extent, entity coreference across documents in a corpus), we found that existing methods did not meet our accuracy requirements when applied to historical documents. 
Some widely used methods require the entity to be in the knowledgebase, whereas historical documents are replete with individuals not remembered in Wikipedia. Indeed, one motivation for disambiguating entities in historical texts is to understand why some people are remembered and others are not. 
Historical texts have a different distribution of entities than modern texts; for example, there may be fewer hyperlinks in crawl corpora - a common source of training data - to these individuals' Wikipedia pages.
Moreover, historical texts often have OCR noise, and language can evolve across time. 
To feasibly run entity disambiguation over massive-scale historical texts, the method also needs to be highly computationally efficient, as academic and archival budgets are typically highly constrained, and the amount of potential historical material to disambiguate is vast. 

This study makes three central contributions designed to improve and encourage further research on cross-document coreference resolution and disambiguation of individuals in historical texts: a massive-scale training dataset replete with hard negatives, high quality evaluation data drawn from historical documents, and entity coreference and disambiguation models trained and evaluated on these data.

Our first contribution is to train coreference and disambiguation models, with a focus on historical texts, and news in particular. 
Our aims are: 1) accurate performance, 2) highly scalable, 3) allows out-of-knowledgebase mentions, and 4) trainable with simple recipes and limited compute.

A contrastively trained bi-encoder retrieval architecture, widely used for open-domain retrieval (\textit{e.g.}, \citet{karpukhin2020dense}), is an excellent fit for these requirements. In a contrastively trained bi-encoder, the neural network encodes mentions (in the document corpus or a knowledgebase) that refer to the same entity nearby in embedding space and encodes mentions referring to different entities further apart. At inference time, a given entity mention is disambiguated to the nearest entry of the knowledgebase if their encodings are within some threshold similarity. If the mention encoding is sufficiently dissimilar to all entries in the knowledgebase, the entity is marked as out-of-knowledgebase. Analogously, an arbitrary number of entity mentions across documents in a corpus (\textit{e.g.,}, different newspaper articles) can be coreferenced by clustering their embeddings.

This approach is highly scalable because each entity mention and entry in the knowledgebase is embedded only once, no matter how many entities are to be disambiguated. Moreover, a Facebook AI Similarity Search (FAISS) \cite{johnson2019billion} backend can be used to retrieve the most similar knowledgebase embedding for each mention in the document corpus. FAISS is extremely optimized, scaling up to billion-scale datasets on relatively modest hardware. This architecture also easily handles out-of-knowledge-base entities. Assuming an appropriate training dataset, it is straightforward to train, making it feasible, even on a highly constrained academic compute budget, to update the models as the deep learning literature advances or to tune them for specific document collections. 

A second central contribution of the paper is the creation of a massive-scale dataset for contrastive training of entity coreference and entity disambiguation models. 
\WikiDataset constructs over 190 million entity pairs for contrastive training. 
Positive pairs come from contexts (paragraphs) in Wikipedia that contain hyperlinks to the same page (for coreference), or from a context and the first paragraph of the relevant entity that it links to (for disambiguation). 

We obtain hard negative entity pairs - \textit{e.g.,} pairs that are highly confusable - at scale from Wikipedia disambiguation pages, which list entities that have confusable names or aliases. The hard negative pairs come from contexts that link to different pages on a given disambiguation page. For example, the disambiguation page "John Kennedy" includes John F. Kennedy the president, John Kennedy (Louisiana politician), John F. Kennedy Jr, and a variety of other John Kennedys. Hard negatives sample contexts mentioning John F. Kennedy (\textit{e.g.,} with hyperlinks to John F. Kennedy's page) and pair them with contexts mentioning other entities from the John Kennedy disambiguation page.
We overrepresent hard negatives from within families (\textit{e.g.}, paired mentions of Henry Ford Jr. and Henry Ford Sr.) by mining family members from Wikidata pages. These challenging cases are very common in historical texts (\textit{e.g.,} fathers and sons with the same name and profession).   
Constructing our open-source \WikiDataset required substantial wrangling of Wikipedia dumps, and we have made it publicly available (CC-BY) to allow other researchers to more easily exploit this rich source of information.  

The extensive hard negative pairs in \WikiDataset allow us to contrastively train our entity coreference and disambiguation models on four Nvidia A6000 GPU cards, a modest setup by deep learning standards. In contrast, contrastive training with random negative pairs requires massive batch sizes to achieve strong performance \cite{he2020momentum}.

We train our \CorefModel coreference model on paired contexts around Wikipedia mention hyperlinks.
We further tune this coreference model for disambiguation, creating \DisambigModel, by training on paired contexts and first paragraphs. If desired, this model can be further tuned on target data. We do so for historical newspapers, creating the \DisambigNewsModel model by tuning on a hand-labeled dataset linking individuals in newspapers to Wikipedia. 

A third contribution is the development of a high quality benchmark that coreferences and disambiguates individuals in historical newswire articles from the 1950s and 1960s, that appear in the Newswire dataset \cite{silcock2024newswire}. We hand disambiguate entities - or mark them as not in the knowledgebase - creating the \EoTU historical benchmark.  

We document performance on \EoTU that significantly exceeds that of other widely used entity disambiguation models. 
We are also competitive in disambiguating individuals in modern benchmarks, especially on the MSNBC and ACE2004 benchmarks (both of which disambiguate modern news), where we outperform other state-of-the-art models.
This suggests that our models and training data have broader applications beyond historical documents, especially to modern news. 

Cross-document coreference is moreover highly accurate. Cross-document coreference resolution is often central to cataloging historical document collections and can be applied as a first step to creating a knowledgebase when the individuals in a corpus are not covered in existing knowledgebases. 

Finally, we briefly illustrate some of the facts that can be gleaned from tagging unique individuals in historical document collections, by applying the \DisambigNewsModel model to a large-scale historical news dataset. 
We are enthusiastic about the promise of coreferenced and disambiguated historical documents to lead to many informative insights. 
Our datasets and models are open-source, with a CC-BY license, and we hope that they encourage further engagement with historical cross-document coreference resolution and disambiguation. 

The rest of this study is organized as follows: 
Section \ref{lit} discusses the related literature, and
Section \ref{dataset} introduces the novel massive scale \WikiDataset training dataset and historical benchmark.
Section  \ref{methods} develops models for entity coreference and disambiguation, and 
Section \ref{eval} evaluates model performance. Section \ref{applications} applies our models to a massive-scale historical news corpus.
Finally, Section \ref{limits} considers limitations, and Section \ref{ethics} discusses ethical considerations.

\section{Literature} \label{lit}

Entity disambiguation has inspired a variety of architectures, including a masked language model (LUKE) \cite{yamada2022global} and a neural translation model (GENRE) \cite{de2020autoregressive}.
While these architectures work well for some problems, they are not well-suited to disambiguation of large-scale historical corpora. 
The masked language model approach limits to the top 50K Wikipedia entries, many of whom are not people, due to computational constraints in computing the softmax. It also does not allow out-of-knowledgebase entities and requires sparse entity priors. The neural translation approach's sequence to sequence architecture is slow at inference time, requiring around 60 times longer to run than other models considered in our comparisons. 

This paper follows the most scaleable entity disambiguation approaches in employing a bi-encoder architecture.
One of the inspirations for the current study is BLINK
\cite{wu2019scalable}, which models entity disambiguation as a text retrieval problem, using a contrastively trained BERT \cite{devlin2018bert} bi-encoder and a re-ranking cross-encoder.
The model assumes all entities are in the knowledgebase.
This study updates the bi-encoder architecture with advances made over the past five years (such as using mean pooling rather than a [CLS] token for the representation \cite{reimers2019sentence} and advances in training efficiency). It also develops an expansive training dataset, replete with hard negatives. We develop a coreference model and incorporate it into the disambiguation pipeline and allow for out-of-knowledgebase individuals. 

Another well-known model that uses a bi-encoder is ReFinED \cite{ayoola-etal-2022-refined}, an entity linking model that performs mention detection and disambiguation for all mentions within a document in a single pass. Like BLINK, it uses a bi-encoder architecture but allows entities to be out of knowledgebase. This study compares the performance of our models to GENRE, BLINK, and ReFinED, widely used models with well-maintained codebases. 

The cross-document coreference resolution literature is less dense, but a closely related study is \citet{hsu2022contrastive}, which uses a contrastively trained RoBERTa \cite{liu2019roberta} bi-encoder and clustering for cross-document resolution of entities and events. They do not consider disambiguation to an external knowledgebase. 

A key distinction between this study and most existing entity disambiguation benchmarks is its focus on real-world contexts with lesser known entities - many lost to history except in the contexts of the documents being considered.
Most benchmarks only contain in-knowledgebase entities. 
An exception is \citet{kassner2022edin}. They detect out-of-knowledgebase mentions by clustering representations, with a cluster defined as out-of-knowledgebase if there are no Wikipedia embeddings in the cluster. They then add the mean embeddings of the out-of-knowledgebase clusters to the knowledgebase index and run entity linking with this fully comprehensive embedding index. They use BLINK as the encoder. To create a dataset with out-of-knowledgebase entities, they link a large crawl corpus (OSCAR) to two Wikipedia dumps taken at time $t_0$ and $t_1$. Links to pages added between $t_0$ and $t_1$ are then out of knowledgebase when disambiguating to the knowledgebase in $t_0$. This type of out-of-knowledgebase entity is different than the type we encounter in historical texts, as they are entities that were prominent enough at the time of mention to link to Wikipedia but aren't in an earlier snapshot. In contrast, in historical applications, there are many individuals who are simply not very prominent. We do not compare on this benchmark, because all these entities were in the late-2022 Wikipedia snapshot we used for training. Hence, they were seen in training by our models (but not by some of the older comparisons) and no longer approximate truly out-of-knowledgebase entities.


\section{Datasets} \label{dataset}

\subsection{Training Data}

High quality hard negatives, as well as paired positive data, are needed to train contrastive models for entity coreference and disambiguation. We create a novel, massive-scale training dataset - \WikiDataset - by mining entity pairs from Wikipedia disambiguation pages and Wikidata family relationships. 

Entity contexts are drawn from a Wikipedia XML dump\footnote{https://dumps.wikimedia.org/} from November 11, 2022, with mentions of each entity appearing as a hyperlink to their page. We then split the entities into train, test, and validation sets, pairing mentions of the same entity along with their context (defined by the paragraph containing the entity mention) to create positive pairs. We create `easy negatives' by pairing an entity mention with mentions of a different entity.
We obtain `hard' negatives using Wikipedia's disambiguation pages, \textit{e.g.}, John Fitzgerald Kennedy and John Kennedy (Louisiana senator).  
We further enhance our training data with in-context negatives, other entities that appear in the context window of the entity under consideration. The resulting dataset, totaling around 185 million pairs, is described in Table \ref{tab:entity_data}.

\begin{table}[ht]
\resizebox{\linewidth}{!}{\begin{tabular}{@{}lccc@{}}
\hline
& Train & Val & Test  \\
\hline
\texttt{Wiki Coreference} & 179,069,981 & 5,819,525 & 5,132,56 \\
\texttt{Wiki Disambiguation} & 4,202,145 & 522,385 & 528,709 \\
\NewsHNDataset & 5046 & 666 & 666 \\
\hline
\end{tabular}}
\caption{\raggedright Statistics on dataset size.}
\label{tab:entity_data}
\end{table}

We also create data for disambiguation by linking contexts with entity mentions to their associated template, forming positive pairs. To create the template, we obtain from Wikidata names, aliases, and occupations/positions held by individuals. For example, for President Kennedy: "John F. Kennedy is of type human. Also known as Kennedy, Jack Kennedy, President Kennedy, John Fitzgerald Kennedy, J. F. Kennedy, JFK, John Kennedy, John Fitzgerald `Jack' Kennedy, and JF Kennedy. Has worked as a politician, journalist, and statesperson." We then append this text with the first paragraph of the associated Wikipedia page.

Easy negatives are created by linking contexts with random entity templates. Similar to our coreference training, we use Wikipedia disambiguation pages to associate entity contexts with hard negative templates. We also create negative pairs using family relationships in Wikidata - \textit{e.g.}, John F. Kennedy and Jacqueline Kennedy Onassis (who like many women in historical documents could often be referred to as "Mrs. John Kennedy"). We split the entities into an 80-10-10 train-validation-test split.

Finally, we further adapt the training domain to newspapers. An advantage of our bi-encoder architecture is that it is straightforward to tune to specific domains, plausibly helpful given historical document collections can be quite idiosyncratic. 

We prepare a hand labeled dataset, \NewsHNDataset to tune our \DisambigModel model to the historical news domain. First, we obtain names and aliases of individuals from Wikidata, then do a sparse search for them in a newspaper corpus spanning a century. We hand label whether the article refers to the anchor (\textit{e.g.,} John F. Kennedy) or someone with the same name or alias (\textit{e.g.,} City Councilman Jack Kennedy). When they refer to different individuals, these form hard negatives. We create extra hard negatives by matching an individual with another individual mentioned in the same context, and Wikipedia hard negatives by matching an individual with another individual mentioned in the same Wikipedia disambiguation dictionary. Easy negatives are created by matching with a random individual. 

\subsection{Evaluation Data}
An important contribution of the study is to create high quality evaluation data for entity coreference resolution and disambiguation with historical documents. 
Our \EoTU benchmark labels historical, off-copyright U.S. news wire articles \cite{silcock2024newswire}. 
We double label 157 news wire articles, from 4 different days from 4 years in the 1950s and 60s, totalling 1,137 person mentions. 
The articles were labeled by highly motivated North American undergraduate students\footnote{They were paid at the rate set by our department.} and all discrepancies were resolved by hand.
We label days on which State of the Union addresses took place, as there are modestly more coreferences to resolve, providing more power for evaluating this task. 

We split \EoTU into a 50-50 val-test split, so the coreference clustering threshold can be chosen on the val split.
Table \ref{tab:benchmark_mentions} compares the size of the \EoTU test split to other widely used modern benchmarks for entity disambiguation. 

\begin{table}[h!]
\centering
\resizebox{\linewidth}{!}{\begin{tabular}{lcccc}
\toprule
\textbf{Benchmark} & \textbf{Total} & \textbf{Mentions in} & \textbf{People} & \textbf{People in} \\
& Mentions & Wikipedia & Mentions & Wikipedia \\
\midrule
AIDA-CoNLL (test) & 1,824 & 1,824 & 248 & 248 \\
ACE2004 & 257 & 257 & 20 & 20 \\
AQAINT & 727 & 727 & 64 & 64 \\
MSNBC & 656 & 656 & 228 & 228 \\
WNED-WIKI & 6,821 & 6,821 & 625 & 625 \\
WNED-CWEB & 11,154 & 11,154 & 1,361 & 1,361 \\
EotU (test) & 569 & 446 & 569 & 446 \\
\bottomrule
\end{tabular}}
\caption{Entity and people mentions across different benchmarks.}
\label{tab:benchmark_mentions}
\end{table}

Note that our full dataset is twice the size of the test set described here. The larger WNED datasets are generated automatically from webtexts - e.g., using links from elsewhere on the web to Wikipedia - and hence are more akin to \WikiDataset. 

Note that these widely used benchmarks have all entities in the knowledgebase. 
In \EoTU (validation and test splits), there are 220 unique individuals that are in Wikipedia, totaling 898 mentions. There are 239 mentions that are not in Wikipedia. 


\section{Methods} \label{methods}

\begin{figure*}[ht]
    \centering
    \includegraphics[width=.9\textwidth]{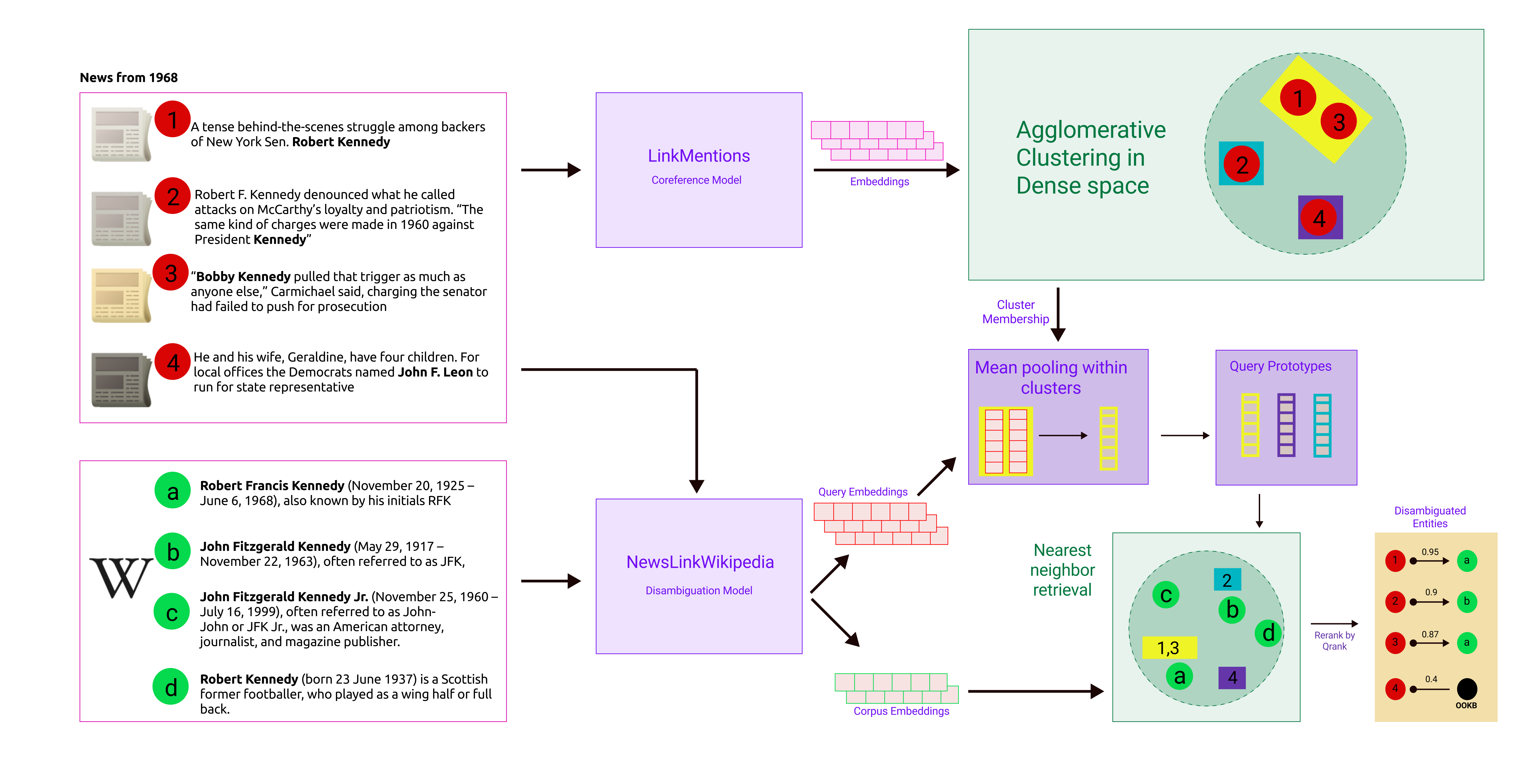}
    \caption{\raggedright Model architecture of \DisambigModel.}
    \label{fig:arch}
\end{figure*}

\CorefModel and \DisambigModel coreference mentions across documents in a corpus and disambiguate person mentions to a knowledgebase, respectively. 
An overview of the model architecture is shown in Figure \ref{fig:arch}. 

We separate named entity recognition - which tags the tokens in a text that refer to named entities - from entity disambiguation, rather than doing them end-to-end as in entity linking. Even in noisy historical news articles, we are able to achieve 94\% accuracy tagging people with named entity recognition as a token classification task. Errors tend to occur when OCR noise is so severe that there is little hope of disambiguating the entity. Hence, there is not much scope for errors to propagate. Separating named entity recognition and coreference/disambiguation simplifies the architecture, making it easier for the social science community to implement or customize to individualized applications. 

We moreover focus on [PER] (person) tags from named entity recognition, as these are of primary interest for many historical applications. We found that locations could be disambiguated very well using non-neural methods and Geonames, a larger structured database of georeferenced locations. 

\textbf{Coreference Resolution Model:} 
Our coreference resolution model links mentions of a given person across documents in a corpus.
Depending on the question at hand, coreference resolution can produce the final output or can be used to create a prototype entity for disambiguation to an external knowledgebase.

To contrastively train \CorefModel, using the 179,069,981 coreference training pairs in our novel \WikiDataset coreference training set, we employ Online Contrastive Loss as implemented in \citet{reimers2019sentence}, with cosine similarity and margin of 0.4, and utilize AdamW as the optimizer with a linear warmup scheduler set to 18.2\%. Our training setup includes 4 Nvidia A6000 GPUs, a batch size of 512, and a learning rate of 1e-5. We train for a single epoch, processing each pair in the training split only once. The best model is chosen based on the pair-wise classification F1 score on the validation set, with the highest validation F1 being 92.75\%.

We use a sequence length of 256, and initialize with an \textit{all-mpnet-base-v2} Sentence-BERT model \cite{reimers2019sentence} from the Hugging Face hub. This is a lightweight model, making it more feasible for those in the academic community with limited resources to train and deploy it at scale. The model is trained in Pytorch \cite{pytorch_lib} with hyperparameters tuned using hyperband implemented in Weights and Biases \cite{wandb_sweep}.
Given the large training dataset, we found it beneficial to divide it into 10 chunks before training. After completing each chunk (1/10 of an epoch), we resumed training from an intermediate checkpoint and lowered the learning rate to 2e-6 after the first chunk to minimize the risk of the optimizer overshooting the minima. Since training each chunk began with a warmup, our approach effectively simulates a linear scheduler with restarts.

We observed significant performance improvement by using special tokens ($[M]$ Entity $[\backslash M]$) around an entity mention \cite{wu2019scalable}. For instance, "Eisenhower sharing a light moment with President-elect $[M]$ John F. Kennedy $[\backslash M]$ during their meeting in the Oval Office." 

At inference time, the contrastively trained \CorefModel is used to embed the mentions, and then they are grouped together via hierarchical agglomerative clustering (with average linkage) using cosine similarity. A threshold of 0.15 was chosen on the validation set. (Other clustering methods are straightforward to swap in, as desired.)
In the newspaper corpus, we coreference entities across articles within dates.

\textbf{Disambiguation:}
For disambiguation, we fine-tune our coreference model on the disambiguation portion of \WikiDataset, with similar hyperparameters to those used in coreference training, but without restarts or chunking. The learning rate is 2e-6, with a 20\% warmup, and batch size is 256. The model is trained for three epochs, and the best checkpoint is selected based on the classification F1, achieving a maximum validation F1 of 97\%.

This model can be further tuned for a specific application. We create \DisambigNewsModel by tuning on the paired disambiguation data in \NewsHNDataset. We use an identical training setup, achieving a maximum validation F1 of 85\%.

Next, we prepare a lookup corpus to disambiguate entity mentions to the correct entity using semantic information from both the context around the mention and information from a template we create from Wikipedia and Wikidata pages of individuals, as described above. We prune our knowledgebase to remove extraneous entities. We include only entities of instance type human, who have a birth or death date, as we found most that didn't were instance type errors. 
We remove pages of individuals born after the conclusion of the corpus and remove entities with no overlap and a high edit distance between the Wikidata label and the associated Wikipedia page title. We found that those pages overwhelmingly were not a person page for that Wikidata entry. 

The resulting knowledgebase has 1.12 million person pages. 
We embed these templates using our disambiguation model and store them in a FAISS IndexFlatIP index \cite{johnson2019billion}.

\begin{table*}[ht]
\centering
\resizebox{\linewidth}{!}{\begin{tabular}{lccccccc}
\toprule
\textbf{Benchmark} & \DisambigModel & \DisambigNewsModel & BLINK & GENRE & GENRE & ReFinED & ReFinED \\
& & & & BLINK Data & AIDA-CoNLL & Base & AIDA-CoNLL \\
\midrule
EotU (all) & 74.0 & 78.3 & 59.9 & 63.4 & 62.4 & 65.4 & 64.0 \\
EotU (in KB) & 85.2 & 89.0 & 76.5 & 80.9 & 79.6 & 60.3 & 59.9 \\
AIDA-CoNLL (test set) & 70.6 & 71.7 & 79.0 & 58.1 & 62.1 & 99.2 & 99.2 \\
ACE2004 & 90.0 & 90.0 & 85.0 & 80.0 & 80.0 & 80.0 & 80.0 \\
AQAINT & 92.2 & 95.3 & 98.4 & 95.3 & 95.3 & 93.8 & 93.8 \\
MSNBC & 89.4 & 98.2 & 81.6 & 82.5 & 84.2 & 84.2 & 82.9 \\
WNED-WIKI & 88.9 & 88.5 & 93.9 & 94.2 & 93.4 & 95.5 & 94.7 \\
WNED-CWEB & 70.7 & 71.5 & 69.1 & 70.7 & 71.5 & 73.3 & 72.4 \\
\bottomrule
\end{tabular}}
\caption{Benchmark performance comparison across different methods. The first row evaluates on all entities in \EoTU, whereas the second row only considers in-knowledgebase entities.}
\label{tab:benchmark_performance}
\end{table*}

To run disambiguation, we embed mentions using the disambiguation model. Using the clusters obtained from coreferencing, we mean pool within each cluster to create the entity prototype embeddings, and use these to query the nearest neighbor(s) in the knowledgebase. 
If there is no embedding in the knowledgebase within a threshold cosine similarity of the query - where this threshold is chosen on a validation set - we mark the entity as not in the knowledgebase. 
If the nearest neighbor is within the threshold distance for a match, and the second nearest neighbor is at least 0.01 cosine distance from the closest neighbor, we disambiguate to the closest neighbor. Otherwise, we utilize Qrank\footnote{\url{https://github.com/brawer/wikidata-qrank/tree/main}}, which ranks Wikidata entities by aggregating page views on Wikipedia, Wikispecies, Wikibooks, Wikiquote, and other Wikimedia projects.
We keep the nearest neighbors that are within 0.01 cosine distance of the closest neighbor to the query. We then we-rank these using Qrank, effectively disambiguating to the most popular entity when the returned matches are very close to each other.

We do not add a re-ranking step with a cross-encoder, as in \citet{wu2019scalable}, in order to maximize scaleability to massive datasets. However, training a cross-encoder on \WikiDataset would be straightforward.

To choose the no match threshold, we annotate the output of our disambiguation pipeline on a set of 6,425 pairs sampled from 13 years in a large-scale newswire dataset \cite{silcock2024newswire}. We then find the cut-off threshold that maximizes pair-wise classification precision and use it as the no-match threshold.

For reproducibility and ease of access, we have made our models and training/evaluation data available on the Hugging Face hub (links are redacted to maintain anonymity for review). All code is in our github repository.

\section{Evaluation} \label{eval}

To measure performance on coreferencing and disambiguating individuals in historical documents, we apply our models and others from the literature to \EoTU. 
We made significant efforts to run existing models on our evaluation data, but not all models in the literature have maintained their codebases, or even made code available. 
Moreover, some models are simply not suitable for the task. For example, LUKE \cite{yamada2019global} limits to top 50K Wikipedia entities (many of whom are not people), meaning relatively few entities in our datasets are in the knowledgebase. We run BLINK \cite{wu2019scalable}, GENRE \cite{de2020autoregressive}, and ReFinED \cite{ayoola-etal-2022-refined}, all prominent models in the entity disambiguation literature with well-maintained codebases. We closely follow their implementations, providing details in the supplementary materials. The latter two models have a zero-shot version and an AIDA-CoNLL fine-tuned version. We report results for both.  

These are disambiguation models, not coreference models. The coreference literature is  thinner and disambiguation is our main focus, and so we do not have comparisons for this task.
Coreferencing individuals across newswire articles is very accurate, achieving an adjusted rand index (ARI) of 96.42. 

Table \ref{tab:benchmark_performance} documents that on the \EoTU dataset, both our zero-shot \DisambigModel model and fine-tuned \DisambigNewsModel model beat other models, by a wide margin. 
\DisambigNewsModel correctly retrieves or classifies as out-of-knowledgebase 78\% of individual mentions, whereas \DisambigModel has an accuracy of 74\%. The next best alternative is ReFinED, which correctly disambiguates around 65\% of mentions. When only considering entities in Wikipedia, \DisambigNewsModel correctly disambiguates 89\% of entities, \DisambigModel correctly disambiguates 85\% of entities, and the next best alternative is GENRE, with an accuracy of 81\%. Hence, while much of the advantage is with out-of-knowledgebase entities, our models also do better on entities in knowledgebase, even zero-shot. 

We also compare performance on disambiguating people in existing, widely used benchmarks. While modern data are not our main focus, our models do reasonably well. In particular, \DisambigNewsModel achieves a near-perfect 98\% accuracy on MSNBC. \DisambigModel has 89\% accuracy, as compared to the next best (GENRE) with 84\% accuracy. This suggests that our models - beyond being suited to historical applications - can also be well-suited to disambiguating modern news. We also beat other models on the news dataset ACE2004, which has very few people. On other modern benchmarks, there are model(s) that perform better but our performance is in the range of the other models.

\begin{table}[h!]
\centering
\resizebox{\linewidth}{!}{\begin{tabular}{lccccc}
\toprule
\textbf{Model} & \textbf{Base} & \textbf{Our Model} & \textbf{Add} & \textbf{Add Birth} & \textbf{Add Qrank} \\
& \textbf{MPNet} & \textbf{Only Disambig.} & \textbf{Coref.} & \textbf{Date Filter} & \textbf{Rerank} \\
\midrule
LinkWikipedia & 26.5 & 62.6 & 73.8 & 73.8 & 74.0 \\
NewsLinkWikipedia & 26.5 & 69.1 & 77.9 & 77.9 & 78.3 \\
\bottomrule
\end{tabular}}
\caption{Ablations. The first column uses a Sentence-BERT MPNet model, which we use to initialize training. Second column discards the coreferencing step, as well as birth date filtering and Qrank re-ranking. The next three columns add back coreference resolution, birth date filtering, and Qrank re-ranking, respectively.}
\label{tab:ablations}
\end{table}

We conduct ablations in Table \ref{tab:ablations}, to quantify the contributions of different elements in our disambiguation pipeline. 
Base MPNet disambiguates with a Sentence-BERT \cite{reimers2019sentence} MPNet \cite{song2020mpnet} model (\textit{all-mpnet-base-v2}), the base model that we initialize with. This model is not intended for entity disambiguation, but we include it to quantify how much is gained through our training. Performance is very poor. The next column reports results from our trained disambiguation models, without coreference resolution or additional processing steps: filtering entities in the knowledgebase to be born prior to the end date of the corpus and reranking by Wikipeda QRank when the nearest entities are very close to each other. Accuracy falls relative to the baseline by around ten percentage points in both models. Adding coreference resolution restores almost all of this decline, with birth date filtering and Qrank reranking contributing little. Coreference resolution plausibly combines information across mentions and reduces noise, leading to overall better quality disambiguation. 

\section{Exploring Entities in Historical News} \label{applications}

To give a flavor of how our framework can be combined with historical documents, we apply our disambiguation pipeline to a large-scale corpus of historical newswire articles - sent out over newswires such as the Associated Press between 1878 and 1977 \cite{silcock2024newswire}. The datatset contains 2.7 million unique articles, reproduced in a corpus of local news over 32 million times. 

\begin{figure}
    \centering
    \includegraphics[width=\linewidth]{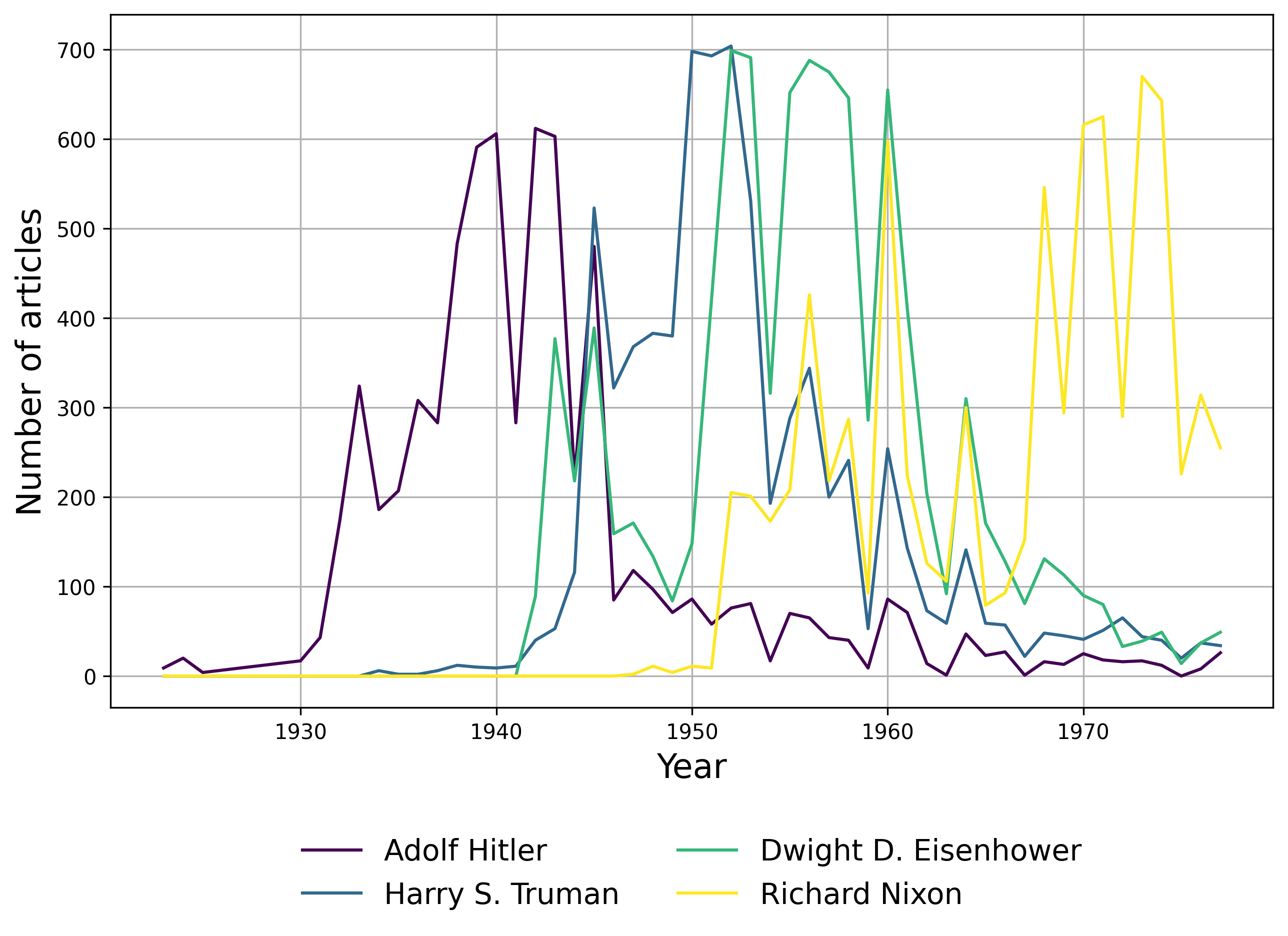}
    \caption{Mentions over time of entities that appeared most commonly in newswire articles.}
    \label{fig:newswire_ents}
\end{figure}

We disambiguate 15,323,463 person mentions, encompassing 61,933 unique individuals. Only 4.6\% of disambiguated entity mentions refer to women, with Golda Meir being the most mentioned woman. The most mentioned entity is Dwight D. Eisenhower, appearing in 9,530 unique articles which are reproduced an average of 33.7 times. Richard Nixon, Harry S. Truman, and Adolf Hitler are the next most mentioned in unique news wire articles. 
Figure \ref{fig:newswire_ents} plots their mention counts. For U.S. presidents, electoral cycles are clearly visible.

Entity disambiguation also allows us to see individuals' occupations in Wikidata. The most common occupations of disambiguated entities are politician, military officer, lawyer, diplomat, military personnel, journalist, trade unionist, actor, economist, businessperson, judge, soldier, baseball player, writer, aircraft pilot, banker, jurist, entrepreneur, and miner.

We plot log mentions by decile in this corpus - a measure of how prominent individuals were in the news historically - against log Wikipedia Qrank by decile, showing the correlation between these two measures. 

\begin{figure}
    \centering
    \includegraphics[width=\linewidth]{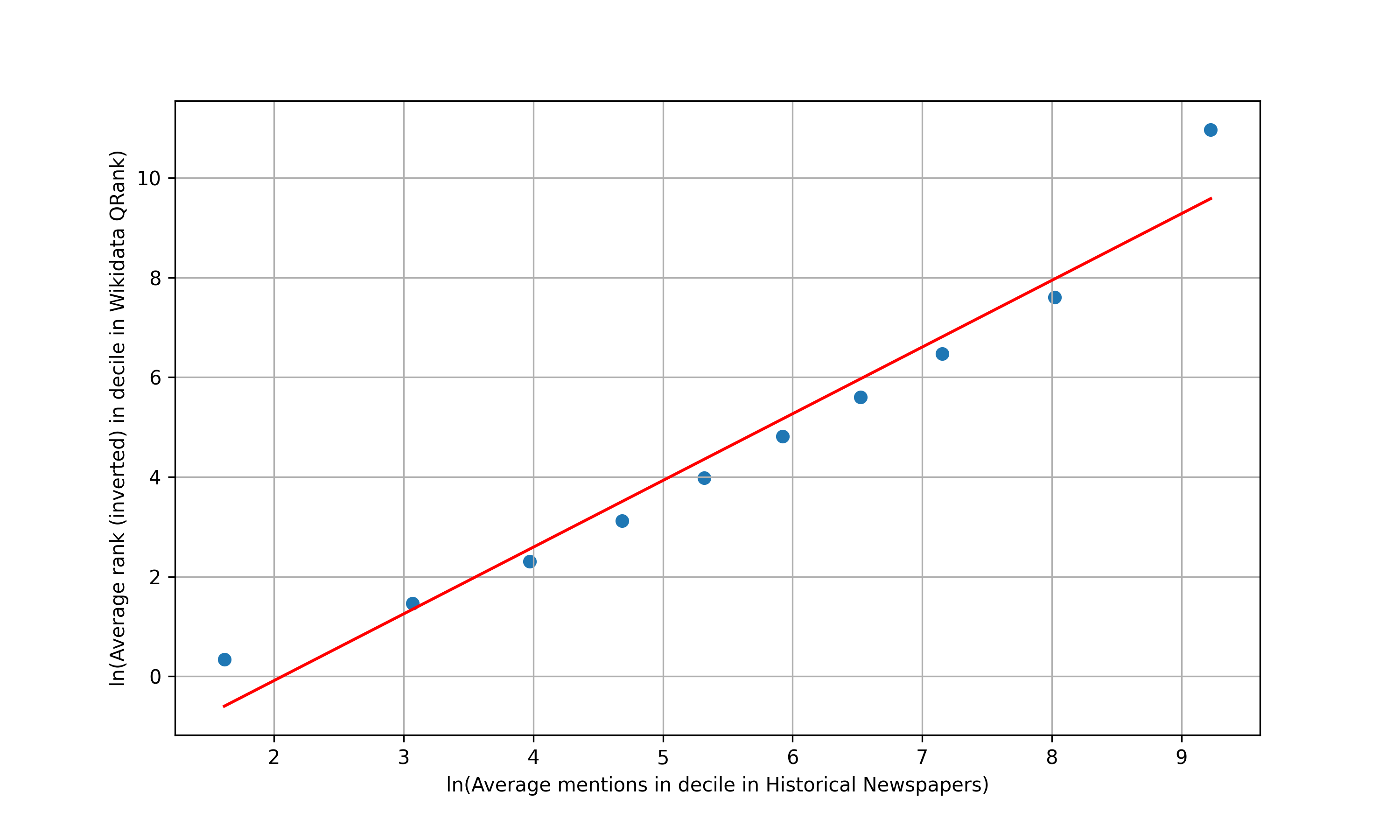}
    \caption{Mentions against Wikipedia Qrank.}
    \label{fig:qrank}
\end{figure}

This dataset, also publicly available, provides fascinating data that researchers can use to study who appeared in historical news and which of these individuals are remembered via Wikipedia today.



\clearpage
\section{Limitations} \label{limits}
The present paper focuses on individuals and disambiguation to Wikipedia/Wikidata. 
We found disambiguating locations to Wikipedia to be unproductive, as we could achieve very strong performance using sparse methods and Geonames, a more comprehensive database of locations. In the future, we hope to extend the model to organizations (though in practice, many historical organizations end up being out-of-knowledgebase).

While Wikipedia is extensive, there are many people who never entered this knowledgebase, and various biases may influence which historical figures are remembered in Wikipedia. Part of the objective of applying \DisambigNewsModel to a massive scale corpus of historical newswires is to understand more about which individuals were considered broadly newsworthy at the time but have since been forgotten. This information could be used to expand databases such as Wikipedia in the future. Nevertheless, we cannot disambiguate an individual who is not in the knowledgebase, no matter how noteworthy they were historically. 

\section{Ethical Considerations} \label{ethics}

This study presents no major ethical concerns. Its
methods are entirely open source, and its training
data are entirely in the public domain. 
We disambiguate individuals in widely reproduced historical newspaper articles, which are in the public domain and hence do not pose privacy concerns. 

It is possible that some applications could raise concerns. Historical news, government publications, and other historical documents reflect the biases of their time and may contain factual inaccuracies or offensive content.
Moreover, while our models are reasonably accurate, they are not perfect, and depending on the usage of the output, human revision of the match - potentially bringing in additional information - may be required.
It is important to interpret the disambiguated texts critically, as is the norm in rigorous historical research.

\clearpage


\begin{LARGE}
    Supplemental Materials
\end{LARGE}

\section{Model hyperparameters}

\subsection{Entity coreference}

At training time, we split the large training set into 10 chunks and we use the following hyperparameters: 

\begin{itemize}
    \item Online contrastive loss, with cosine distance metric and a margin of 0.4
    \item Batch size: 256
    \item Max sequence length: 256
    \item Epochs: 1
    \item Learning rate 1e-5, increased to 2e-06 after first chunk
    \item Optimiser: AdamW 
    \item Warm-up rate: 0.182, increased to 1 after the first chunk
\end{itemize}

At inference time, we cluster entities using Hierarchical Agglomerative clustering, with average linking and cosine distance metric, with a threshold of 0.15, which was validated on the validation split.  

\subsection{Entity disambiguation}

At training time we use the following hyperparameters:

\begin{itemize}
    \item Online contrastive loss, with cosine distance metric and a margin of 0.4
    \item Batch size: 256
    \item Max sequence length: 256
    \item Epochs: 3
    \item Learning rate 2e-6
    \item Optimiser: AdamW 
    \item Warm-up rate: 0.2
\end{itemize}

\section{Comparison to other entity disambiguation models}

We compare our method against existing entity disambiguation architectures. Results from this are given in the text. Details of the implementations of other architectures follow.

\subsection{BLINK}

To run BLINK \citep{wu2019scalable}, we use the implementation from their github repository directly (\url{https://github.com/facebookresearch/BLINK}). Specifically we use \verb|main_dense.run()| with default hyperparameters. 

\subsection{GENRE}

For GENRE \citep{de2020autoregressive}, we used their huggingface implementation (\url{https://huggingface.co/facebook/genre-linking-blink}). We used 10 beams, with truncation around the entities to a max token length of  384 as suggested in the appendix of the original paper. Otherwise default hyperparameters were used.  We evaluate using both their base model (\verb|facebook/genre-linking-blink|) and their model finetuned on Aida-YAGO.

\subsection{ReFinED}
To evaluate ReFinED, we use entity disambiguation implementation from their github repository (\url{https://github.com/amazon-science/ReFinED}). We use \verb|refined.process_text()| with default hyperparameters. For each mention, we specify that the mention type is PER using \verb|coarse_mention_type = "PERSON"|.

\section{Annotator instructions}

\subsection{\NewsHNDataset annotator instructions}

\NewsHNDataset annotators were given the following instructions: 

\begin{itemize}
    \item ``The box at the top shows the entity in question, the years in which they held some kind of important position, and their aliases (see more detail at the end of this email). 
    \item It also shows a code starting with Q. This is the unique reference for the person in Wikidata. You can query search this there (https://www.wikidata.org/) and pull up more detail about the person, including information on positions they've held, what else they've done etc. Most of these people also appear on Wikipedia, so this is also a good source for finding out more info about the person. 
    \item Then there are (up to) 32 passages of text. In each of these there is a highlighted term. Your job is to label each of these passages for whether the highlighted term is the same as the person in the box at the top (`positive') or not (`negative'). In some cases this will be pretty clear, and in some cases it might need a bit of digging. 
    \item In some cases we don't have 32 passages, so you'll see some turn up as "Empty" at the end - no need to label these ''

\end{itemize}

\subsection{\EoTU annotator instructions}

To create \EoTU, 1,137 entity mentions across 157 newswire articles were double-annotated by undergraduate research assistants. 
Annotator labeling instructions were as follows:

``We've pulled out lots of articles from the day of the State of the Union speech in 1958, 1959, 1960 and 1961. A team of RAs over this semester has been labelling the spans in these texts that refer to an entity. What we want you to focus on is working out which unique entity these spans refer to. The database of unique identifiers that we will work from is Wikidata (https://www.wikidata.org/). If you search for entities on here, you will see they have a unique identifier beginning with Q (eg. [example redacted to preserve anonymity of authors]
). For each entity in the articles we'd like you to find the unique identifier in wikidata (if it exists). It might also be useful to use wikipedia in difficult cases - all pages on wikipedia will have an entry in wikidata, but wikidata is bigger than wikipedia, so there might be some entities in wikidata that you don't find in wikipedia [...] there's 14 entities to label on average per article. The entity in question should be highlighted in the text.''

Annotators were encouraged to reach out with questions and clarifications.


\bibliography{cites}
\end{document}